# Knowledge-Informed Machine Learning for Cancer Diagnosis and Prognosis: A review

Lingchao Mao[1,∧], Hairong Wang[1,∧], Leland S. Hu[3, 4, 5, 6], Nhan L Tran[5, 6], Peter D Canoll[7], Kristin R Swanson[2, 5], Jing Li[1*]

[1] H. Milton Stewart School of Industrial and Systems Engineering, Georgia Institute of Technology, Atlanta, GA, 30332, USA.
[2] Mathematical NeuroOncology Lab, Precision Neurotherapeutics Innovation Program, Mayo Clinic Arizona, 5777 East Mayo Blvd, Support Services Building Suite 2-700, Phoenix, AZ, 85054, USA.
[3] Department of Radiology, Mayo Clinic Arizona, 5777 E. Mayo Blvd, Phoenix, AZ, 85054, USA.
[4] School of Computing, Informatics, and Decision Systems Engineering, Arizona State University, 699 S Mill Ave, Tempe, AZ, 85281, USA.
[5] Department of Neurosurgery, Mayo Clinic Arizona, 5777 E. Mayo Blvd, Phoenix, AZ, 85054, USA.
[6] Department of Cancer Biology, Mayo Clinic Arizona, 5777 E. Mayo Blvd, Phoenix, AZ, 85054, USA.
[7] Department of Pathology and Cell Biology, Columbia University Medical Center, 630 West 168th Street, New York, NY, 10032, USA.

∧Co-first author
*To whom correspondence should be addressed: jli3175@gatech.edu

**Abstract**

Cancer remains one of the most challenging diseases to treat in the medical field. Machine learning has enabled in-depth analysis of rich multi-omics profiles and medical imaging for cancer diagnosis and prognosis. Despite these advancements, machine learning models face challenges stemming from limited labeled sample sizes, the intricate interplay of high-dimensionality data types, the inherent heterogeneity observed among patients and within tumors, and concerns about interpretability and consistency with existing biomedical knowledge. One approach to surmount these challenges is to integrate biomedical knowledge into data-driven models, which has proven potential to improve the accuracy, robustness, and interpretability of model results. Here, we review the state-of-the-art machine learning studies that adopted the fusion of biomedical knowledge and data, termed *knowledge-informed machine learning*, for cancer diagnosis and prognosis. Emphasizing the properties inherent in four primary data types including clinical, imaging, molecular, and treatment data, we highlight modeling considerations relevant to these contexts. We provide an overview of diverse forms of knowledge representation and current strategies of knowledge integration into machine learning pipelines with concrete examples. We conclude the review article by discussing future directions to advance cancer research through knowledge-informed machine learning.

# Introduction

Cancer stands as a predominant cause of human death worldwide, with its incidence escalating alongside the increasing global life expectancy[2]. Despite its widespread occurrence, cancer remains one of the most formidable challenges in the medical field because the genetic and pathogenic mechanisms of tumors are still too complex to be fully resolved. In recent decades, machine learning (ML) has positioned itself as promising tool for analyzing complex patterns from large datasets. The computational power and versatility of ML models allow for the discrimination of subtle differences in multimodal images, analysis of vast arrays of digital histopathology slides, and interpretation of complex genetic and molecular profiles. ML has demonstrated success in numerous cancer applications and has promising potential to automate medical data analyses, improve diagnosis and prognosis capabilities, and support better clinical decision-making in cancer treatment[3,4].

A leading cause of the limited effectiveness of conventional therapies is the pronounced heterogeneity inherent in tumors[5–8]. At the individual-level, no two patients, even within the same subtype of tumor, behave clinically the same, with or without treatment, suggesting that conventional one-model-fits-all approaches are not sufficient. At the tumor level, genomics studies reveal spatial heterogeneity within tumors[9–12], underscoring the presence of distinct cellular subpopulations with varying phenotypic features even within the same sample of a primary tumor, a phenomenon known as *intratumoral heterogeneity*. Moreover, the application of treatments like chemotherapy imposes selective pressure on tumor cells, resulting in treatment-resistance cells dominating the tumor mass and driving disease progression, thereby reshaping the tumor landscape[5]. Considering these intricacies, there is a pressing need to develop accurate and effective models that can delineate the spatial landscape within tumors and facilitate targeted cancer treatment.

Generalization performance of machine learning models hinges on the availability of both high-quality and enough training and test data. Nevertheless, the data labeling process usually requires meticulous case-by-case examination by trained medical experts. In practice, the acquisition of large amounts of tumor specimens for hispathological evaluation is infeasible, as only a few biopsies can typically be obtained from restricted tumor locations of a patient. These constraints hinder the ability of machine learning models alone to learn the complete spatial landscape of tumors. Yet, we know that there is wide interpatient and intratumoral heterogeneities within and across tumors that urgently need to be predicted for individual patients to not only inform diagnosis but also to allow for prediction and tracking of treatment response.

Another important modeling challenge in cancer applications is how to analyze diverse, multimodal, high-dimensional data in a methodical manner to provide clinical predictions. Recent studies highlight the potential of integrating individual genomic phenotypes with medical images, a field known as *radiogenomics*, as a promising alternative to invasive diagnosis[13]. However, the typical number of patient samples is on the order of hundreds, while the human genome comprises tens of thousands of genes. Machine learning confronts the challenge of finding relevant patterns from a multitude of genetic features mixed with measurement noise, irrelevant genes, and inherent biological variability[14]. Developing models that can effectively integrate multimodal and high-dimensional data under small sample sizes is an important direction for enhancing AI-assisted cancer prognosis.

Lastly, a lack of interpretability can undermine trust in machine learning models as decision-support tools and limit the clinical actionability of the predictions. In particular, deep learning models often face criticism for being "black-box machine learners", with mechanisms that are unintelligible and unverifiable to humans. Explainable AI (XAI) techniques have emerged to address this concern, aiming to quantify the importance of features in model's decision-making process[15]. While these interpretation tools allow practitioners to validate model behavior against medical knowledge, they are generally applied post-hoc. A more effective approach is to integrate the expected behavior into the model training process, a topic we delve into in the next section.

To address the aforementioned challenges, a viable strategy involves the integration of biomedical knowledge into ML models, referred to as *knowledge-informed machine learning (KIML)*. By regularizing the model's learning process using domain knowledge (or models of that knowledge), the accuracy, robustness, and interpretability of models can be improved. Over the past decade, KIML has garnered increasing interest and demonstrated success across scientific, engineering, and health applications, particularly as a solution for settings with limited training data.

Several reviews have delved into the concept of knowledge-informed machine learning. Karpatne et al.[19] were pioneers in defining the concept of theory-guided data science, providing a review of ML models fused with theory-based models in scientific applications. Alber et al.[20] conducted a review encompassing methods that combine ML with ordinary and partial differential equations, focusing on applications in biological, biomedical, and behavioral sciences. More recently, Kim et al.[21] surveyed methods for integrating prior knowledge into deep learning (DL) models, specifically for modeling dynamical systems. The first comprehensive taxonomy of KIML was proposed by von Rueden et al.[22], however, limited application examples were provided.

This review focuses on the application of KIML in the cancer domain, where rich biomedical knowledge exists to navigate the aforementioned modeling challenges. The scope encompasses both traditional ML and DL models utilized for cancer diagnosis and prognosis, employing input data (X) such as gene expression profiles or medical images to predict clinical outcomes (y), such as cancer phenotype or patient risk score. An extensive search was conducted on PubMed in April 2023, focusing on original research articles published from 2012 onward (within the last 10 years). The search was employed via the concatenation of three keywords: (*knowledge-informed, physics-informed, knowledge, informed, theory-guided, mathematical model*) and *(machine learning, deep learning)* and *cancer*. We acknowledge that this search criteria might not cover all important work in the field. A total of 127 articles met the inclusion criteria for this review (Figure 1).

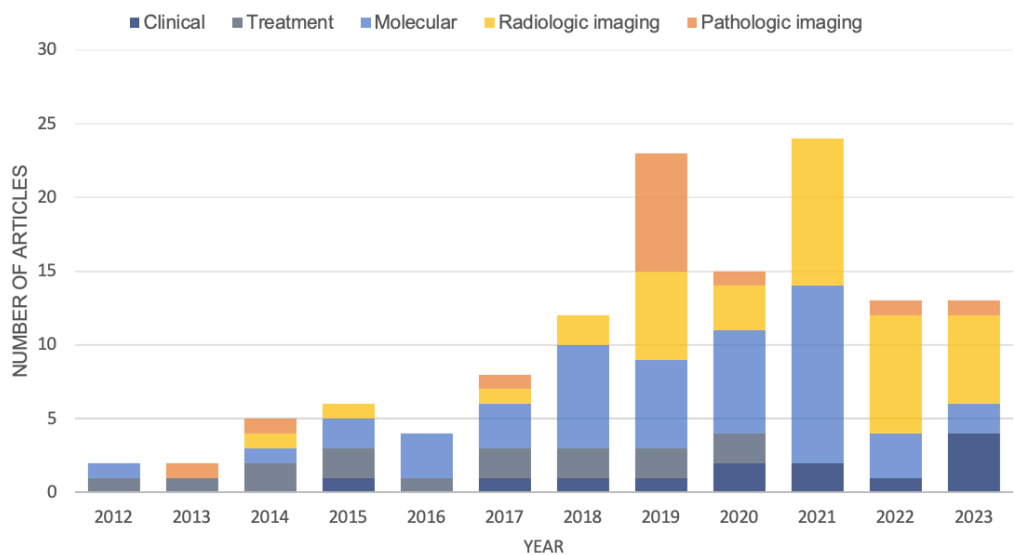

Figure 1. **Annual trends of knowledge-informed machine learning articles addressing cancer diagnosis and prognosis.** The total number of articles included in the analysis was 127, spanning from January 2012 to April 2023. Articles that involve multiple data types are singularly counted based on one of their associated data types.

In this review, we organize our discussion around three components, as illustrated in Figure 2. We begin by describing the four major types of data pivotal for predictive modeling of cancer. We then scrutinize diverse forms of representing biomedical knowledge. Subsequently, we provide a comprehensive overview of mainstream strategies for integrating knowledge across various stages of machine learning, including data preparation, feature engineering, framework design, and model training. Finally, we discuss future directions aimed at advancing KIML for accurate predictions of cancer diagnosis, prognosis and response to treatment.

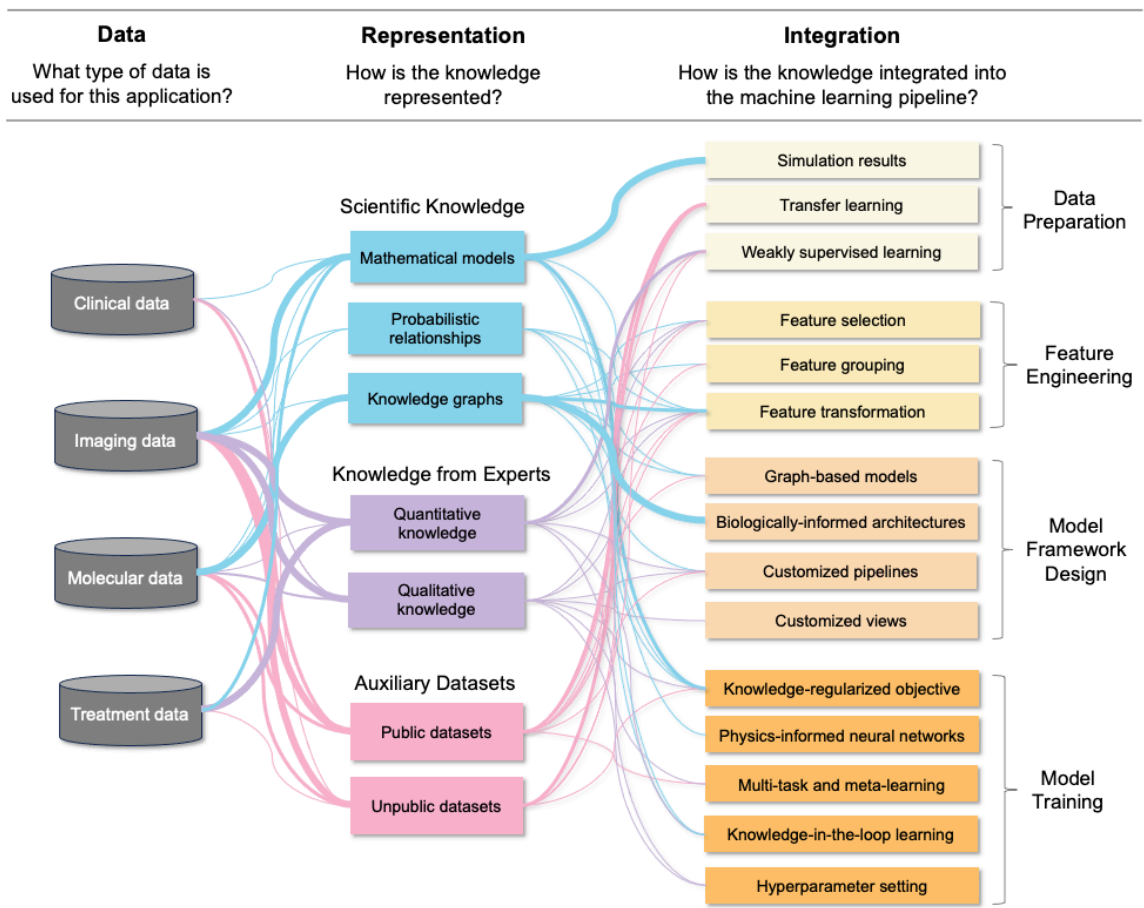

Figure 2. **Taxonomy of knowledge-informed machine learning in cancer diagnosis and prognosis**. Our literature review categorizes existing along three dimensions: type of data, form of knowledge representation, and strategy for knowledge integration. Note that one paper may be included in more than one category. The thickness of the paths indicates the relative frequency of papers in each area (thin: one to four papers; medium: five to nine papers; thick: equal or more than ten papers).

## Types of input data

A foundational review[23] defined four types of features that should be considered when developing a prediction model for clinical decision-support systems (CDSSs). Adhering this framework, we describe four primary types of data in cancer diagnosis and prognosis: clinical, imaging, molecular, and treatment data. An overview of their capabilities and limitations is provided in Table 1.

Table 1. **Summary of main cancer datasets with their capabilities (C) and limitations (L).**

| Type of data | Description | Capabilities | Limitations |
|---|---|---|---|
| Clinical | Contain a spectrum of patient-related information, including demographics, medical history, laboratory test results, symptom logs, performance status, questionnaires, health records, and more | (C1) Diverse perspectives of patient's health condition (C2) Relatively inexpensive to collect; often collected as part of the standard clinical procedure | (L1) Unstructured and heterogeneous (L2) Limited standardization standards across medical centers (L3) Textual data require models that can handle sequential data, thus not straightforward to fuse with non-textual data modalities |
| Radiologic Imaging | Medical examinations done through imaging technologies such as MRI, PET, CT, X-rays, ultrasound, and mammography | (C1) Non-invasive way to localize the lesion (C2) Provide high-level morphologic and phenotypic measurements about the tumor and its surrounding tissue/host organ (C3) Existence of commonly used standardization and preprocessing protocols | (L1) Acquisition is expensive and requires equipment; each imaging modality comes with additional acquisition cost (L2) Resolution and noise level depend on the imaging technology (L3) Presence of artifacts (L4) Inter-observer variability |
| Pathologic Imaging | Image taken from tissue samples obtained from patients through biopsy procedure | (C1) Provide low-level characterization of specific cell-level features of tissues (C2) Allow finding abnormalities that point to disease's upstream causes | (L1) Invasive and expensive to acquire (L2) Constraints in data collection such as tissue sample location and quantity (L3) Time-consuming labeling and shortage of pathologists with expertise to annotate datasets (L4) Inter-observer variability (L5) Large size images demanding more memory and computational resources |
| Molecular | Describe the abundance or status of molecules in tissue samples. Multi-omics data span genomics, epigenomics, proteomics, transcriptomics, etc. Molecular interactions data describe potential function of molecules through their interactions with other partners. | (C1) Provide low-level characterization of specific molecular-level features of tissues (C2) Multiple assays can be conducted for each tissue sample (C3) Provide insights into tumor heterogeneity and progression (C4) Single-cell technologies facilitates generation of this data | (L1) Invasive and expensive to acquire (L2) Constraints in data collection such as tissue sample location and quantity (L3) High-dimensional outputs with high sparsity, measurement variability, and noise |
| Treatment | Information about cancer treatment plans such as radiotherapy or chemotherapy doses and medication usage. | (C1) Important for monitoring and understanding post-treatment tumor progression (C2) Relatively cheap to collect; generally already available | (L1) Generally reported as mean dose; spatial distribution of doses not always available |

MRI, Magnetic Resonance Imaging; PET, Positron Emission Tomography; CT, Computer Tomographic

## Clinical data

Clinical data encompass a spectrum of patient-related information, spanning demographics, medical history, laboratory test results, symptom logs, performance status, questionnaires, health records, and more. Typically, clinical data can be obtained with relative ease[23], though the use of validated scoring systems is imperative for ensuring standardized clinical measures across medical centers. A large portion of clinical data is provided in text form, such as electronic health records (EHR), nursing notes, pathology reports, radiology reports, and more. These diverse data sources offer diverse perspectives on the patients' health condition.

With the emergence of successful language models in the deep learning field, there is a growing research trend to employ deep learning techniques for the automatic annotation of clinical notes with disease outcomes such as cancer metastasis, disease stage, or tumor recurrence. In particular, Large Language Models such as BERT[24], BioBERT[25], BioMegatron[26], and clinicalBERT[27] have demonstrated promising potential for automated knowledge extraction from textual data. Pre-trained on large biomedical text corpora such as PubMed, these models can support clinicians in the evaluation of their findings in comparison to existing evidence in the literature[28]. It is worth noting that text data is typically analyzed separately from other data modalities and using different model architectures due to inherent differences in data format and information density. The coordination of unstructured textual data with other data types such as imaging is still an active area of research.

## Imaging data

### Radiologic imaging

Radiologic imaging constitutes a routine component in the clinical diagnosis, surveillance, and treatment monitoring of tumors. Common types of radiologic imaging employed in cancer applications include Magnetic Resonance Imaging (MRI), Positron Emission Tomography (PET), Computer Tomographic (CT) scans, X-rays, ultrasound, and mammography. Radiologic imaging allows capturing pathophysiological and morphological characteristics of the tumor in a noninvasive manner. Despite these advantages, radiologic images can be limited by the spatial resolution and noise of the imaging technology. The presence of artifacts stemming from factors such as movement, scatter, attenuation, and partial volume effects (PVEs) can further impact imaging quality, potentially leading to ambiguous tumor boundaries[29].

In the clinical domain, the diagnosis of suspicious lesions as benign or malignant rely on the visual interpretation and expertise of radiologists, introducing potential inter-observer variability[30,31]. Computer-aided diagnosis systems allow for more reproducible descriptors through systematic processing of quantitative tumor features[32]. Recently, ML has gained widespread application in analyzing radiologic images, facilitating automated tumor segmentation, diagnosis, and monitoring. These AI-powered systems hold promising potential for expediting the time-consuming human examination process. A prevalent approach involves building whole-tumor classification models, generating a singular predicted outcome for each tumor[33]. However, given the intratumoral heterogeneity, a spatially resolved assessment of tumor landscape is needed to provide better therapeutic value. As demonstrated by a recent study[34], radiomics models capable of delivering predictions across each spatial unit of the tumor can be used to identify tumor regions with genomics alterations.

Another pertinent research question is how to integrate multi-modal images with missing data. Consider multiparametric MRI as an example. Each imaging sequence unveils distinct characteristics of the tissue: T1-weighted sequences reflect blood-brain barrier and regional angiogenesis; T2-weighted sequences can be used to assess extracellular fluid in brain parenchyma; and Diffusion tensor imaging (DTI) aid determining the water diffusion process, which is affected by tumor cell architecture and density[35]. Although multi-modal imaging data affords a more comprehensive understanding of the tumor and surrounding abnormal tissues[36,37], the collection of complete multi-modal data for every patient may not be feasible due to economic or practical constraints. How to leverage incomplete-modality data and maintain prediction performance for patients who only possess a subset of modalities is an active area of research.

Additionally, imaging techniques such as MRI and CT suffer from long acquisition times, which leads to patient discomfort, motion artifacts, and expensive acquisition cost. Substantial effort has been made to accelerate MRI acquisition by only collecting subsampled k-space data, then use ML to reconstruct the high-resolution image. Traditional methods based on low-rank matrix completion or compressive sensing assume sparsity or low total variation to regularize the ill-posed reconstruction problem. Recently, deep learning methods that incorporate prior knowledge about imaging physics demonstrated promising potential for this reconstruction task[38,39].

### Pathologic imaging

In contrast to radiologic images that provide high-level characterization of the tumor, pathologic images focus on sparsely sampled cell-level evaluations for each localized tissue. While radiology has a long history of research and clinical application, pathology just underwent a digital revolution after Food and Drug Administration's approval of digitized whole-slide images in 2017[40]. With enhanced storage and data management capabilities, digital pathology slides have proliferated as an indispensable instrument in clinical diagnostics, drug development, tissue-centric research, and other scientific endeavors.

The relative novelty of digital pathology and high cost of manual microscopic evaluation are reasons for the lack of large, annotated datasets. As pathologists are trained to follow algorithmic decision trees to assess the specimens, notable *inter-observer variability* exist[41]. A retrospective study[42] found that pathology review by experts changed the nodal status in 24% of patients. Another challenge is the intrinsic high-resolution nature of whole-side images (WSI) rendering them too large for certain ML models to accommodate under memory and computational constraints. The prevailing approach is to partition WSIs into non-overlapping or sliding window patches for model training, then aggregate patch-level predictions to form slide-level diagnosis. However, the patch-based approach entails a compromise on the broader visual context, potentially resulting in an incomplete representation of information crucial for diagnosing specific phenotypes[43]. An additional stratum of challenge arises from the frequent unavailability of region-specific annotations that precisely identify tumor indicators, instead of image-level annotations, owing to the time-intensive labeling process. There is no recognized best strategy for how to map image-level labels to patch-level partitions and how to aggregate patch-level results into image-level diagnosis, as no single rule adequately encapsulates the diagnostic behavior exhibited by pathologists.

### Molecular data

Cancer is essentially a disease of the genome which evolves with accumulations of somatic mutations. Tailoring molecularly targeted treatments based on genetic changes offers a personalized approach to enhance clinical outcomes beyond conventional one-size-fits-all treatments. Genomics data plays a pivotal role in disease diagnosis, drug development, and the identification of biomarkers and immune signatures. For instance, cancer patients harboring specific genetic alterations in the EGFR gene can receive treatments precisely targeting those changes.

Advances in gene sequencing and computational tools have enabled the comprehensive study of the entire genome, encompassing genomic (DNA mutations), epigenomic (chromatin or DNA states), proteomic, transcriptomic, metabolomics, and other -omic levels[44]. In essence, *molecular omics data* characterize the abundance of molecules in tissue samples, providing intricate insights into cancer clonal evolutions, tumor heterogeneities, and cell radiosensitivity at a much higher resolution[23]. Another type is *molecular interaction data*, which describe the potential function of molecules through interacting with others[44], such as protein-protein interactions and protein-RNA interactions. Various molecular interaction datasets, derived from large-scale cell line studies, are publicly available[45–52].

A benchmarking review[53] of state-of-the-art predictive models at the NCI-DREAM7 Drug Sensitivity Prediction Challenge found that models using all omics profiles (genomic, epigenomic, proteomic) exhibited the best performance for drug response prediction in breast cancer cell lines, suggesting that different omics profiles provide complementary information for the predictive task. The authors also emphasized the importance of incorporating prior biological knowledge regarding breast cancer oncogenes and disease-driving pathways[53].

Despite the inherent strengths of molecular measurements, the acquisition of tumor specimens is inevitably expensive. Obtaining biopsy samples necessitates surgery and subsequent preprocessing for sequencing tests, and due to practical constraints, only a limited number of biopsy samples can be collected from restricted locations. This leaves a substantial portion of the tumor region unmeasured. The emergence of liquid biopsy, which collected through a peripheral blood draw, offers a less invasive alternative. However, not all clinically relevant biomarkers can be detected from liquid biopsy, and many tests derived from this venue require further clinical validation[54,55]. Moreover, conducting molecular tests on every tumor is sometimes prohibitive, as treating a patient with multiple drugs to study the independent effects of each drug is impractical. A viable alternative is to conduct pre-clinical experiments using human cancer cell lines or animal models. Assuming that genetic associations remain consistent to some extent between cell line and patient tumor samples, larger pre-clinical datasets provide valuable insights for clinical models of drug response prediction[56,57]. It is important to note that trials of molecular biomarkers are prone to experimental variability[23], thus standardizing assay criteria or harmonizing multi-omics data is imperative before incorporating molecular features into predictive modeling.

### Treatment data

Conventional cancer therapies involve surgery, chemotherapy, and radiotherapy. Surgery that majorly removes tumor is the most effective treatment at early stages, however, the possibility of tumor recurrence is high[58]. While chemotherapy and radiotherapy can be used before/after surgery to eliminate the remaining tumor, tumor cells may become resistant to the chemotherapy drug[58]. Features extracted from planned spatial and temporal distribution of the radiotherapy or chemotherapy dose are valuable information for predicting tumor recurrence and post-treatment progression[23]. Various studies have integrated treatment plans, dose reconstructions, or historical clinical drug doses as either predictive features or ground truth labels in ML models[59–62,62–72]. Salient works utilized patient-specific anatomy features and planning parameters in their dose prediction models[59,63,71]. Typically, doses are reported in terms of mean dose applied to the entire tumor or doses to a prescription point inside the tumor. Nevertheless, recording spatially variable dose distributions can provide more value for models that consider intratumoral heterogeneity[73]. In clinical practice, the interplay of treatment time and radiotherapy doses is recognized to impact tumor progression[23].

## Types of knowledge representation

The form in which knowledge is encoded shapes how it can be integrated into ML models. In the subsequent subsections, we describe diverse forms of representing scientific knowledge, expert knowledge, and knowledge present in auxiliary datasets. Table 2 provides an overview of knowledge representations.

Table 2. **Summary of forms of knowledge representation and their capabilities (C), limitations (L), and integration strategies (M)**

| Representation | Description | Capabilities and Limitations | Integration strategy |
|---|---|---|---|
| Scientific Knowledge | | | |
| Mathematical Models | Knowledge expressed through mathematical expressions that contain variables or constants.<br><br>Examples: PI model of glioblastomas[1], model of immune system-tumor interactions[74] | (C1) Well-defined and mathematically-expressed<br>(C2) Interpretability and transparency<br>(C3) Generalization for being extensively validated through scientific experiments<br>(L1) Limited expressiveness and accuracy, may not capture complex relationships or may not account for all factors<br>(L2) Computationally expensive and slow to run if the equation has no closed form solution<br>(L3) Limited availability | (M1) Simulation as training data[29,58,75–81]<br>(M2) Knowledge-regularized objective[78,79,82–88]<br>(M3) Physics-informed neural networks[83,89,90]<br>(M4) Feature transformation[80,81]<br>(M5) Knowledge-in-the-loop learning[58,75,77] |
| Probabilistic Relations | Knowledge expressed as connections between variables in a probabilistic or stochastic setting, such as correlations or probability distributions.<br><br>Examples: gene mutation rates, probability atlas | (C1) Contains uncertainty quantification of the knowledge so can be more robust to noisy data<br>(C2) Easy to incorporate as prior in Bayesian models<br>(L1) Relatively imprecise and weak form of supervision<br>(L2) Certain degree of subjectivity | (M1) Feature transformation[91]<br>(M2) Knowledge-in-the-loop learning[92,93]<br>(M3) Knowledge-regularized objective[94,95]<br>(M4) Hyperparameter setting1/12/24 2:00:00 AM |
| Knowledge Graphs | Knowledge expressed as a graph where vertices represent concepts or biological entities and edges denote relationships between them. | (C1) Structured representation facilitates explainability of complex relationships and cross-domain insights<br>(C2) Several large-scale knowledge-graphs are publicly accessible | (M1) Knowledge-regularized objective[94,97–100]<br>(M2) Biologically-informed neural network architectures[98,101–111] |
| | Examples: gene databases, protein-protein interaction networks, pathway databases | (L1) Limited resources available for non-genetic data<br>(L2) Incomplete information, gaps in knowledge may limit the model performance and may discourage exploration of unknown relationships.<br>(L3) Struggle to integrate with unstructured data | (M3) Feature selection, grouping, and transformation[60,112–126]<br>(M4) Customized model pipelines1/12/24 2:00:00 AM<br>(M5) Graph-based models[56,128–131] |
| Knowledge from experts | | | |
| Quantitative knowledge from experts | Quantitative assessments provided by experts based on their medical experience.<br><br>Examples: tumor location annotations, treatment plan selections, ordinal relationships, lesion size estimates | (C1) Direct feedback highly relevant to the target task<br>(L1) Require time-consuming assessments by experts<br>(L2) Inter-observer variability | (M1) Weakly supervised learning[40,101,132–134,134–137]<br>(M2) Customized model pipelines[38,138,139]<br>(M3) Hyperparameter setting[138,140,141]<br>(M4) Knowledge-in-the-loop learning[68,142]<br>(M5) Feature selection and transformation[59,63,64,71,143,144] |
| Qualitative knowledge from experts | Non-numerical information derived from observations, interviews, subjective experiences, and behaviors of experts from their clinical practice.<br><br>Examples: glaze of pathologists when examining an image, lexicon of relevant clinical vocabulary, "rules of thumb" in diagnosis | (C1) High-level and nuanced information about the task<br>(C2) Generally already available with low acquisition costs<br>(L1) Not straightforward to express mathematically or to incorporate into the model, usually requires more customization efforts to tweak model behavior<br>(L2) Not well-defined and subjective<br>(L3) Imprecise supervision | (M1) Feature selection and transformation[145–150]<br>(M2) Customized model pipelines[134,151–153]<br>(M3) Customized model views[154,155]<br>(M4) Knowledge-regularized objective[43,122] |
| Auxiliary datasets | | | |
| Public datasets | Openly accessible and shared collections of data, typically collected from large-scale studies and from multiple centers, spanning various formats such as image, text, omics, dictionaries.<br><br>Examples: TCGA, PubMed, UMLS, KEGG, ImageNet | (C1) Wide accessibility<br>(C2) Available with larger number of samples, various modalities, and different disease conditions<br>(C3) Facilitates benchmarking with existing studies<br>(C4) Transparency<br>(C5) Diverse topics and applications<br>(L1) Data quality variability and heterogeneities in data acquisition protocols, equipment, batch effects, and cohort differences.<br>(L2) Limited specificity, not tailored to specific research questions<br>(L3) Overused or outdated data | (M1) Transfer learning[28,115,145,156–160]<br>(M2) Weakly supervised learning[161,162]<br>(M3) Feature selection, grouping, transformations[14,76,163,164]<br>(M4) Customized model pipelines[165]<br>(M5) Multi-task and meta-learning[166,167]<br>(M6) Graph-based models[168] |
| Private datasets | Datasets that are not shared publicly and access is restricted to authorized individuals or | (C1) Exclusive access to valuable and proprietary information to | (M1) Transfer learning[139,169–173]<br>(M2) Knowledge-regularized objective[56] |

| | entities to protect sensitive patient information and comply with legal and ethical standards.

Examples: pre-clinical data, patients with other types of cancer, database of treatment plans, data from healthy controls | foster unique insights and discoveries<br>(C2) Control over data quality<br>(C3) Tailored to specific research questions<br>(L1) High acquisition cost<br>(L2) Limited collaboration<br>(L3) Challenges in benchmarking | (M3) Feature transformation[61,144,149] |
|---|---|---|---|

PI, Proliferation-Invasion; TCGA, The Cancer Genome Atlas; KEGG, Kyoto Encyclopedia of Genes and Genomes; UMLS, United Medical Language System

## Scientific Knowledge

We categorize knowledge in connection with science, technology, mathematics, or findings from scientific research as scientific knowledge. Knowledge in this category is obtained through a series of observations, phenomena, formulation of hypotheses, and validation using scientific methods. In the following subsections, we discuss three prevalent representations of scientific knowledge, namely mathematical models, probabilistic relations, and knowledge graphs. It is important to note that some works extend beyond these three categories[38,39,33].

## Mathematical Models

Mathematical models, expressed through equations containing variables and constants, distill knowledge derived from extensive experiments and clinical trials. The growing use of mathematical models to integrate and test existing hypotheses and knowledge in cancer biology have driven forward the field of Mathematical Oncology over the last couple decades[174]. The advantage of employing this form of knowledge lies in its well-defined nature and extensive validation through experiments like those using foundational mathematical biology approaches summarized in the pioneering J. D. Murray Book "Mathematical Biology" in the context of cancer, leading to a diversity of insights across cancer applications[175,176]. These mathematical models play a pivotal role in integrating and testing hypothesis in cancer research.

Several mathematical models have been established for cancer growth[1,177,178], immune system-tumor interactions[74], normal tissue complications in radiotherapy[179], imaging physics[180,181] or thermal dynamics[89]. For instance, the proliferation-invasion (PI)[1] model is a differential equation that simulates the growth and invasion of glioblastomas (GBM), one of the most aggressive types of brain tumor characterized by diffuse invasion and high recurrence rates:

$$\frac{\partial c}{\partial t} = \nabla \cdot (D \nabla c) + \rho c \left(1 - \frac{c}{K}\right),$$

where $c(x,t)$ is the tumor cell density at location $x$ and time $t$, $D(x)$ is the net rate of diffusion taken to be piecewise constant with different values in gray and white matter (mm$^2$/year), $\rho$ is the net rate of proliferation, and $K$ is the cell carrying capacity. This model has been widely used to estimate intratumoral tumor cell density[182–184], radiation sensitivity[185,186], treatment response[187–190], and gene mutation status[1,79].

Mathematical models and ML serve distinct yet complementary roles in understanding tumor-environment states. The former offers mechanistic understanding of tumor dynamics, while the latter explores data-driven patterns from supplied data. Models should embrace the complementary strengths of these approaches to combine the wealth of known scientific evidence with the ability for large-scale data analysis[16–18]. For example, scalar parameters from mathematical models can be replaced with patient- and/or time- dependent variables estimated from patient-specific datasets using machine learning[84–86,88]. Conversely, mathematical models can be wrapped into the optimization objective of ML models converting them into knowledge-informed loss functions[83,89,90].

Nevertheless, certain mathematical models cannot be easily integrated into ML models due to their intricate formulations, such as those lacking closed-form solutions. To overcome this limitation, simulation proves to be a widely employed numerical method. Typically, a simulation engine employs numerical methods to solve a mathematical model and produces outcomes based on specific parameters within a given context. Example works that included this type of knowledge are a partial differential equation (PDE)-based simulator applied for brain tumor[78,79] and a simulation-based kernel built for kernelized ML models applied for anti-cancer drug sensitivity prediction[80]. Calibration of mathematical models using parameter estimates obtained from experimental data is another venue to reduce the parameter space. For parameters that are difficult to obtain experimentally due to measurement challenges or unintelligible physics of the environment, ML has been a provable venue to estimate suitable parameter values through learning input-output mappings[87,88,177].

## Probabilistic Relations

Probabilistic relations describe connections between variables in a probabilistic or stochastic setting. Knowledge in this context can be represented through correlation structures, probability distributions, or conditional independence relationships between random variables. Examples of integrating this type of knowledge into ML models include utilizing the differential dose-volume histograms (DVHs) distribution in predicting radiation therapy outcome[59,82], incorporating gene mutation rates for different cancer types derived from previous studies to enhance cancer prognosis[91], utilizing a probability atlas containing prior probability of certain organs appearing at each pixel location to aid tumor segmentation[92,93,95], and leveraging zonal probabilities that indicate where aggressive cancer is more likely to occur as the prior in Bayesian models[96].

A representative example is the mathematical framework for DVHs predictions. DVH[192] is one of the most commonly used metrics in radiation oncology for evaluating the quality of an intensity modulated radiotherapy (IMRT) plan, an advanced type of radiation therapy used for tumor treatment. The treatment target is divided into sub-volume $V(A_{jk})$. For each sub-volume, a differential DVH, $dV(A_{jk})/dD$, is calculated and fitted to a skew-normal probability density function,

$$f(p_1, p_2, p_3; D) = \frac{1}{\pi p_2} \exp\left(-\frac{(D-p_1)^2}{2p_2^2}\right) \times \int_{-\infty}^{\frac{p_3(D-p_1)}{p_2}} \exp\left(-\frac{t^2}{2}\right) dt,$$

where $p_1, p_2, p_3$ are parameters of location, scale, and shape. Such knowledge have been incorporated in deep learning models to improve the quality of treatment plans[59,82].

## Knowledge Graphs

Graphs find extensive application in diverse fields, including computer science, physics, social sciences, and engineering, to model complex systems and analyze relationships between variables. A graph $G$ can be defined as sets $(V, E)$, where $V$ is a set of vertices and $E$ is a set of edges connecting the vertices. Vertices typically correspond to concepts, while edges denote abstract relationships between them.

Relationships between gene expressions and higher-level biological entities can be naturally represented as knowledge graphs. Commonly used publicly available knowledge graphs include the Kyoto Encyclopedia of Genes and Genomes (KEGG)[45] database with interaction probability of pairs of genes and proteins, protein–protein interaction database (STRING)[47], molecular signatures database (MSigDB)[49], Gene Ontology[48], gene connection and function prediction (GeneMANIA)[52], human tissue-specific networks[51], the human cellular signaling pathway interaction database (PID)[50], and the Reactome[46] pathway database. For example, the KEGG[45] offers comprehensive insights into the diverse functionalities and applications of biological systems, spanning from individual cells to complete organisms and ecosystems. As one of the most popularly used biological databases, KEGG has been integrated into ML models to enhance model performance across various tasks such as prediction of survival of lymphoma and ovarian cancer patients[100], lung cancer patients[115,117], and GBM patients[114], diagnosis of the grade of Low-Grade Glioma and GBM tumor samples[116], and prognosis of metastasis for breast cancer patients[193].

## Expert Knowledge

Expert knowledge, in our framework, is defined as knowledge held within a specific group of experts and cultivated within their community. Certain aspects of this expertise are extensively studied and documented, while other facets are directly derived from domain experts, implicitly validated through their cumulative experiences.

### Quantitative knowledge from experts

Quantitative knowledge refers to information that employs numbers, measurements, statistics, logical relationships, and other quantitative methods to articulate connections between distinct variables or quantities within a system or process. This category of knowledge is typically well-defined and can be mathematically expressed, making their integration into ML models convenient. While these type of feedback from experts is typically directly related to the target task, their acquisition require time-consuming assessments.

A primary category of quantitative knowledge from experts encompasses diagnoses and annotations. Several works have developed knowledge-based treatment planning models by leveraging physician satisfied treatment plans or clinical plan doses for prostate cancer[59,61,63,64,67,71,194], head and neck cancer[62,68,68,70–72] and oropharyngeal cancer[65,66]. A representative example by Chanyavanich et al.[61] assembled a database of a hundred prostate cancer IMRT treatment plans, all reviewed and delivered by experts. Each new case was matched to a prior reference case within the knowledge base based on the relative spatial locations of target volume and normal structures. The treatment plan parameters from the matched case were used as an enhanced starting point in the planning process. Similarly, to diagnose future occurrences of Cytokine Release Syndrome, a common adverse effects of chimeric antigen receptor therapies in cancer treatment, Bogatu et al.[137] defined a knowledge base (KB) of statistical biomedical facts extracted from domain literature. The KB was used to augment samples of biomarker concentration measurements through quantifying the degree of similarity between the given measurement value and the statistical values reported in each study in the domain literature[137]. Beyond the diagnosis, experts' estimations of disease progression offer valuable supplementary information. In a detection and diagnosis model of lung cancer, malignancy scores estimated by radiologists were employed in model training as imprecise labels[132]. Further, coarse annotations or drawings of abnormal regions by pathologists have been utilized to aid cancer type classification and image segmentation[133,140,141].

Experts may provide knowledge in the form of ordering relationships, hierarchical relationships, or correlations between entities. A paradigm of integration of ordinal relationships is provided by the work of Mao et al.[195] in brain tumor prognosis, where the authors utilized the ordinal relationship concerning tumor cell density in samples from different areas to extract pairs of ordered samples for model training. A similar strategy was adopted by Wang et al.[135] to model knowledge-determined ordinal relationships between labels for genetic status prediction. On the other hand, Hu et al.[139] observed a consistent positive correlation between relative cerebral blood volume (rCBV) and tumor cell density and leveraged this information to constraining knowledge transfer from other patients.

Another type of quantitative knowledge pertains to estimated sizes or shapes of lesion areas. For example, Zhou et al.[138] noted that tumors generally exhibit the largest aspect ratio in the coronal view, and the shape of tumors in the sagittal view is closer to square than in the other two views. Leveraging this information, appropriate anchor sizes and aspect ratios for each view were specified, forming the basis for constructing candidate bounding boxes for view-specific tumor detection models. A similar example of using expected size and shape distributions of nodules in a deep learning segmentation model was done by Liu and others[152].

## Qualitative knowledge from experts

Qualitative knowledge encompasses information derived from observations, interviews, subjective experiences, and behaviors of experts from their clinical practice, drawing non-numerical connections between different entities within the system. While this form of knowledge is typically available at low acquisition costs, they are not well-defined and are accompanied with subjectivity. Adapting the model behavior to follow such "expert experience" require more customization efforts.

Certain qualitative knowledge about diseases or human biology, while seemingly intuitive to humans, can significantly enhance ML models. For instance, tumors of the same category are often found in similar anatomical locations: gliomas typically involve white matter, meningiomas are commonly adjacent to the skull, gray matter, and cerebrospinal fluid, and pituitary tumors are often located near the sphenoidal sinus, internal carotid arteries, and optic chiasma[35]. Chen et al.[150] proposed a tumor region augmentation and partition strategy to take into consideration the differently weighted informative context surrounding the tumor. Other qualitative aspects such as the smoothness of morphological characteristics of cells within a WSI[134], sparsity of predictive gene expressions[105], transferability of drug sensitivity between cell lines and patient tumor samples[56], boundary ambiguities of the lesion in MRI[59], experts' validation and feedback[196,197], and visual heterogeneities[134,148,148] have been attempted to account for in ML models.

Another qualitative knowledge shared by the medical community are domain-specific lexicons. Liu et al.[145,146] curated a lexicon comprising relevant clinical words and lists of synonyms extracted from a small sample of radiology reports. This lexicon was used to standardize and trim down input tokens into clinically relevant vocabulary. Guan et al.[164], on the other hand, employed the National Library of Medicine[198] to find gene terms associated with lung cancer. Similarly, other researchers have utilized gene functional categories from the Gene Ontology database[48] to find biologically meaningful groupings of genes.

Experts may also serve as behavioral references for ML models[43,147,155]. Liu et al.[152] captured sonographic characteristics that radiologists focus on when examining ultrasound images. The datasets were then categorized into multiple groups based on the sonographic characteristics selected by radiologists. Similarly, Corredor et al.[147] tracked the gaze of pathologists when they used a navigation window to examine WSI images of the skin. A likelihood of cancerous was assigned to each nucleus based on the number of times the region was examined by pathologists. Drawing inspiration from actual clinical decision-making processes, models have been designed to mimic various zoom levels[154] and different visual cues[151] that pathologists use in their diagnosis.

## Auxiliary datasets

Auxiliary datasets have proven valuable in numerous ML works as they can offer additional information not present in the training datasets. Utilizing an auxiliary dataset can yield several advantages, including: (i) *Improved accuracy and generalizability*: the inclusion of data from external sources can broaden the distribution encountered during training, enhancing the model's ability to learn more robust relationships within the data; (ii) *Domain adaptation*: an auxiliary dataset can facilitate the adaptation of a pre-trained model to a new domain or task. For example, a model pre-trained on a large number of natural images can be fine-tuned on a relatively small set of medical images. This transferability can alleviate the sample size needs of data-hungry deep learning models.

### Public datasets
To encourage collaboration in cancer research, there is increasing research efforts to compile and release the data collected from multiple large-scale preclinical and clinical studies, such as the National Cancer Institute's Cancer Research Data Commons (CRDC)[199], the Cancer Cell Line Encyclopedia (CCLE)[200], the Cancer Genome Atlas (TCGA)[201], International Cancer Genome Consortium[202], Genomics of Drug Sensitivity in Cancer[203], and the Cancer Dependency Map (DepMap)[204]. These public datasets typically encompass large patient cohorts with different cancer types and conditions to serve diverse research topics and applications. The availability of public datasets has greatly fostered collaboration, transparency, benchmarking in ML research. Researchers need to be aware of the data quality variability and differences in data acquisition protocols, equipment, batch effects when using public datasets.

Several studies have demonstrated that harnessing large public datasets, even those unrelated to the cancer domain, can enhance model performance on cancer applications. For instance, Vu et al.[157] and Wang et al.[158] utilized ImageNet for pretraining tumor segmentation and tumor classification models. Liu et al.[145] employed the BERT model pretrained on millions of words from the internet as well as Wikipedia articles to augment training data for their cancer classification model from radiology reports. Another interesting use case by Kwon at al.[76] leveraged healthy tissues from the Cancer Imaging Archive (TCIA) to improve generalizability of their tumor segmentation model.

### Unpublic datasets

In most research projects, data is typically gathered and owned by specific organizations with access restricted to authorized individuals or entities. These datasets are common particularly in healthcare, where protecting sensitive patient information and compliance with legal and ethical standards is paramount. This exclusive access allows researchers to have a better control over data quality and tailor the data collection to specific research questions. Representative use cases of these auxiliary datasets include: inclusion of other MRI sequences from the same patient[205], using data from other patients with the same type of cancer for personalized prediction tasks[139,149,170–172], examination of patients with different types of cancer or drugs[115,169], formulating different tasks within the same domain[166], transfer knowledge learnt from pre-clinical data[57], using a comprehensive database that contains progression notes, pathology and radiology reports, nursing notes[206], or leveraging a collection of physician-approved treatment plans[61,144].

In addition to labeled datasets, there may exist unlabeled or baseline datasets from which the model can extract knowledge relevant to the predictive task, such as unlabeled specimens from other patients[156,159] or data from healthy controls[149,173]. An representative example of this approach first extracted a lower-dimensional representation of normal brain appearance using T1 MRI data of healthy controls[149]. Images of patients with brain tumors were then projected into this representation to obtain a reconstruction of their "normal brain." The feature maps of tumor patients and healthy controls were then aligned using a Siamese network, and regions with low feature consistency between the feature maps of normal and tumor images were segmented as tumor regions.

## Types of knowledge integration

In this section, we review how biomedical knowledge can be integrated into ML models. We categorize knowledge integration strategies according to which stage of the ML pipeline the integration happens: data preparation, feature engineering, framework design, and model training.

### Data preparation

Knowledge integration by training data augmentation is an intuitive way to incorporate additional datasets generated based on prior knowledge. Strategies such as transfer learning and weakly supervised learning can leverage both original data and knowledge-generated data for model training.

### Simulation data as training data

Synthetic data generated through simulation can be combined with observed data for model training[29,58,75–81]. Simulation proves valuable for assessing scenarios that may be impractical or costly to explore in physical settings. A representative model is the widely employed computer-based glioma image segmentation and registration (GLISTR) algorithm[76]. GLISTR takes a generative approach to produce patient-specific tumor segmentation maps. The process initiates with a set of normal atlases derived from the healthy population, each defined as a probability map for white matter, gray matter, and cerebrospinal fluid. The atlases are then customized to patient-specific maps by simulating tumor growth using the diffusion-reaction-advection model. This simulation process incorporates tumor shape priors and a user-defined seed radius. The Expectation Maximization algorithm is employed to iteratively estimate the parameters of the tumor growth model and refine the posterior probability of tissue labels.

A recent work employed simulated data for pre-training a segmentation model of PET[29]. The first- and second-order tumor descriptors were extracted to generate simulated images based on the physics of PET. To ensure visual realism, tumor seed locations were manually selected, and intra-tumoral heterogeneity was introduced by incorporating unimodal variability or modeling the intensity as a mixture model. While all simulated images shared the same ground truth as the original tumor image, they featured diverse realistic background intensities from multiple patients to mimic patient heterogeneity. Finally, a deep learning model underwent pre-training on a substantial number of simulated images and subsequent fine-tuning using a smaller set of real PET images.

## Transfer learning

Transfer learning stands out as a widely employed technique for knowledge integration[28,57,102,115,139,139,145,156–160,166,169,170,207,208]. In essence, this ML technique utilizes a pre-trained neural network model as a starting point for learning a new but related task. It is assumed that by pre-training on an extensive and diverse dataset, the model acquires generally useful feature representations and thus only requires minor fine-tuning on a small dataset to perform the target task. Transfer learning proves particularly valuable in scenarios where there is limited labeled data for the target task, and a substantial amount of labeled data is accessible for other related tasks or from a similar domain.

Incorporating the concept of self-supervised learning, which leverages unlabeled data to learn meaningful representations, can further enhance the efficacy of transfer learning. Self-supervised learning techniques, such as contrastive learning or predictive modeling, enable a model to understand and interpret the structure of unlabeled data by predicting missing parts or identifying similarities and differences within the data. This approach can complement traditional transfer learning by providing a richer and more diverse set of feature representations derived from unlabeled datasets, which are often more readily available and scalable. Thus, self-supervised pre-training, followed by transfer learning, offers a robust framework for leveraging both labeled and unlabeled data, significantly enhancing the model's performance, especially in domains where labeled data is scarce or expensive to obtain. In cases where the auxiliary dataset lacks labels, unsupervised models like autoencoders can be employed for pre-training[14,102,156], along with self-supervised techniques to maximize the utility of available data.

Transfer learning enables flexible transfer of knowledge learnt in different domains. Knowledge can be transferred across patients with the same disease, where a model trained on a comprehensive dataset of historical patients captures population-level patterns, and a smaller, patient-specific dataset is used to bias the model for each individual[139,156,172,195]. Similarly, knowledge can be transferred across different diseases. Garcia et al.[115] conducted pre-training of a CNN model using gene expression profiles encompassing over 30 cancer types and fine-tuned the model on data specific to lung cancer patients. Knowledge can even be transferred across different data modalities. For example, a model trained on commonly used imaging modalities (e.g., T1-Gd MRI) can be transferred to modalities with less training data[95,171]. This has proven benefits in medical imaging applications where patients may undergo varying numbers of imaging exams based on factors like physician preference, cost considerations, and center availability. Moreover, models trained for different ML tasks (e.g., tumor detection, tumor classification, tumor segmentation) using the same dataset can leverage shared learned feature representations[166].

Large scale pre-trained language models, including BERT[24], have gained popularity in various domains. Biomedical variations of BERT, such as BioBERT[25] and BioMegatron[26] trained on biomedical research articles, RadBERT[209] trained on radiology reports, and ClinicalBERT[27] on clinical notes, have been developed. These pre-trained models are later fine-tuned using in-domain data, such as electronic health records from a small group of patients[145,207]. Some studies[157,158] even demonstrated value of pre-training CNN models on non-medical images from ImageNet. While pre-training is typically seen in studies that use public large-scale datasets[115,145,156–159], other types of transfer learning such as instance transfer and feature transfer are more used for unpublic datasets[139,169–171].

The extent of knowledge transferred can be adaptive, contingent on the knowledge- or data-driven similarity between domains or patients. An effective transfer learning algorithm should determine which knowledge, and from which source domains or patients, is transferrable to the target domain or patient. A case of controlled knowledge transfer is exemplified by the work of Chanyavanich et al. where the authors compiled a database of physician-approved treatment plans and implemented an algorithm to identify analogous patient cases from the database[61]. Once the optimal match was identified, the clinically approved plan was incorporated to formulate a treatment plan for the new patient. In another example by Hu et al.[139], knowledge transfer was constrained to patients exhibiting a substantial positive threshold between rCBV and tumor cell density to avoid negative transfer in patient-specific transfer learning models.

## Weakly supervised learning

In addition to transfer learning, weakly supervised learning serves as another approach to augment training data and incorporate knowledge[40,132,134–136,148,161,162,195,206,208,210]. Weakly supervised models can accommodate samples with imprecise or incomplete labels, commonly referred to as weakly labeled samples. Particularly suited for scenarios where obtaining fully labeled data is costly or practically challenging, weakly supervised learning has demonstrated efficacy across a spectrum of healthcare applications.

Imprecise labels provide inexact or coarse-grained information about the ground truth. This scenario is common in pathological applications involving WSI. As discussed earlier, WSI slides are often divided into smaller patches for model training, yet only slide-level annotations are available. Imprecise labels indicate that if a slide is positive, at least one patch must contain a tumor, and if a slide is negative, then all tiles must be devoid of tumors. Multi-Instance Learning (MIL) is a prevalent approach to address such challenges[40,132]. The MIL formulation induces learning of a patch-level representation that can separate the discriminative patches in positive slides from all other patches. Another strategy is self-distillation, which separates weakly labeled data into high and low fidelity samples, selecting only the most representative high-fidelity data for noise-robust training[142]. Alternatively, Joshi et al.[210] employed a rule-based heuristic to extract Standard Uptake Values (SUVmax) of metabolically active tumors from radiology reports. The differences in SUVmax values between longitudinal scans were then utilized to derive weak labels for the disease status of each patient, with a Siamese-style network employed to predict disease progression.

Incomplete supervision occurs when only a limited subset of training samples is labeled, leaving the remaining samples unlabeled. Integrating unlabeled samples into the training process can broaden the feature space and enhance the diversity of instances, thereby improving the robustness of the model. Semi-supervised learning strategies are often employed in such scenarios. Even though specific label measurements are absent for each sample, higher-level domain knowledge may exist regarding the unlabeled samples. For instance, Wang et al.[208] utilized the known hierarchical relationship among three gene modules to predict the regional distributions of intratumoral heterogeneity across the entire lesion for each patient. Another form of incomplete supervision involves ordinal relationships among unlabeled samples[136,148,195]. In the work by Mao et al.[195], it was assumed that the boundary of the Enhancing Tumor (ET) area are likely to exhibit higher tumor cell density than the boundary of the Non-Enhancing Tumor (NET) area. Based on this intuition, pairs of weakly labeled samples from radiologist-annotated Regions of Interest (ROI) were extracted as a source of patient-specific data. A similar concept was applied to compare survival times between two patients and encode this ordinal relationship along the tumor samples[148]. Wang et al.[136] proposed a classifier that could leverage interval labels (samples belonging to either class 1 or 2) and imposed a mathematical constraint to guide the model not to classify these samples into class 3, even though their belonging to whether class 1 or 2 is unknown. Weak labels have the attractive advantage of increasing the size of training data without having to going through the time-consuming annotation process.

## Feature engineering

Feature engineering is a critical part of ML involving the transformation of raw input data into a format suitable and effective for model training. Effective feature engineering can significantly boost the performance of a model compared to supplying raw data. A comprehensive simulation study of disease phenotype classification and drug response prediction showed that incorporating biological knowledge about either predictive genes or signals downstream in a pathway (e.g. proteins) can be more useful than providing the complete set of genes and would require significantly less training data than uninformed learning[121]. Thus, domain knowledge can play an important role in feature engineering.

## Feature selection

In applications involving textual data, standardized medical language databases such as the United Medical Language System (UMLS) and Metathesaurus[211] can be used to curate lexicons of clinically relevant terms and their synonyms. When dealing with gene expression data, a common strategy is to select features based on gene sets or pathways known to be associated with the disease of interest[60,122,164]. However, such approaches present certain limitations. First, focusing only on known biological knowledge may restrict the exploration of gene-disease relationships that are yet undiscovered. Second, the disease marker identification may not fully account for higher-order interactions of genes related to the disease.

A more adaptable strategy involves combining knowledge-informed and data-driven feature selection. A comparative study of 14 published methods for patient risk stratification in breast cancer indicated that models incorporating prior knowledge can enhance gene selection stability and biological interpretability[193].

In predicting relapse for patients with breast cancer, Johannes et al.[118] ranked the influence of genes based on protein-protein interaction (PPI) networks and their fold-change information. This ranking was then used to dampen the chance that features are removed in Recursive Feature Elimination (RFE). Another notable approach is the use of a Bayesian model for feature selection[112,143]. In this method, instead of limiting the number of significant genes, the coefficients of each gene are modeled with a Gaussian prior that depends on the covariance matrix across genes and a latent indicator for their inclusion. Insights from previous studies regarding the stability of each gene, interrelationships among genes, and gene groupings can inform the design of this prior. A similar approach has been employed in radiomic feature selection[143] and derivation of specialized lexicons from radiology reports[145,146].

### Feature grouping

Prior knowledge can be used to form biologically meaningful grouping of features. Introducing such a grouping structure serves as guidance for the model to explore complex relationships among features. For instance, genes that are functionally related or belong to the same pathway or protein can be grouped together[86]; in the preprocessing of textual data, related medical terms can be combined[7,62,72,73]; and nearby similar image patches can be grouped into phenotype groups as a way to model spatial heterogeneity[129]. Once groups are defined, a straightforward approach is to extract features at the group level[87]. While pathway-level grouping is the most common and has been shown effective in several genetic studies[7,135], this structure can be extended to multiple levels of hierarchy, such as genes, pathways, subfunctions, and functions. A more sophisticated strategy entails employing multi-view or multi-modality models to learn weights specific to features and views, as well as relationships both within and between groups.

Grouping of gene expressions can be desired for several reasons. First, signature genes for the classification task are not always available across datasets due to different gene annotations across platforms, which impedes transferability. Second, gene expression profiling can be easily affected by nonbiological batch effects coming from the technical platform. Using pathways have been shown to be more reproducible and robust than treating each gene independently[123,212]. Third, employing biologically structured inputs can facilitate model interpretation and the discovery of disease mechanisms.

### Feature transformation

The input features can be reformatted using scientific knowledge to more intuitively connect the inputs and outputs of the predictive task. For example, drug molecule can be represented as a graph following their chemical structure, where nodes are atoms and edges are bonds between atoms[56,113]. Another study argued that it is not straightforward to establish connection between the physicochemical properties of drugs and the cellular mechanisms of their action[124]. Instead, the authors used associated drugs with proteins and created PPI-network based features[124].

With the growing popularity of convolutional networks like U-Net, genetic data has been transformed into "gene expression images" for training CNN models. Studies have attempted to induce meaningful spatial relationship in the transformed image by ordering gene expressions into a square matrix based on their position in a functional hierarchy[115–117], according to chromosome numbers[33], in a swim-lane organization of pathways[114], or by spectral clustering the gene expression matrix[125]. For instance, Ma et al.[116] used the hierarchical gene structure from KEGG[45] to construct a five-layer tree mapping gene expressions based on the child nodes of the tree. The tree is then converted into an image using the pivot method, considering their adjacency in the tree and sorting genes based on median values across all samples. In addition, genetic data can also be transformed into networks. Weiskittel et al.[126] used the network construction algorithm to create regulatory networks for each cancer lineage in DepMap using CCLE RNAseq data. Four categories of network parameters were calculated and used as input to understand if network features can contribute to gene dependency prediction.

Mechanistic models can also be employed to transform input features into a more informed latent informative space, subsequently utilized for outcome prediction[80,81]. A representative example used a mathematical model that accounts for immune system-tumor interactions to extract immunological features, which were then combined with clinical features as inputs into ML models that predict tumor size at different time points[81].

Knowledge-informed transformations of features not only enhance the predictive utility of raw data but also harmonize data collected from different platforms. Gao et al.[123] performed a transformation on the gene expressions of each tumor sample, converting them into a functional spectrum of enrichment scores through Gene Set Enrichment Analysis with respect to a public gene database. These enrichment scores encode transcriptomic patterns previously demonstrated to be associated with biological function. Such transformations can be applied even if some gene expressions are unannotated or missing in one dataset. A similar strategy can be employed for textual data. For instance, Min et al.[163] mapped clinical reports into Concept Unique Identifiers of the UMLS to extract base concepts and hierarchies of related concepts from different reports. These mappings aid in standardizing medical and non-medical terms across diverse data sources.

## Model framework design

To better adapt the natural structure inherent in knowledge, many studies proposed specialized model structures that allow incorporation of biomedical knowledge. Framework design, in this context, refers to the process of designing the modeling pipeline as well as the architecture of backbone models. Deep neural networks, in particular, provide significant flexibility to design different model architectures suitable for different forms of knowledge.

### Graph-based models

Given that cancer is caused by complete processes, it is unlikely that genes act in isolation; rather, they interact with other genes through complex signaling or regulatory networks[168]. Graph-based models provide a natural solution to analyzing both the lower-level and higher-level interactions between biological entities. In a graph, edges can be defined as undirected or directed, deterministic or probabilistic, making them adaptable to different types of knowledge. Edges play a pivotal role in propagating information across connected nodes, a process referred to as *belief propagation*. This propagation property of graph models enables easy extension to semi-supervised learning settings. Information about the outcome can be propagated from labeled patients to unlabeled patients based on the similarity between their features[79,168].

The flexibility of graph-based models has led to extensive use in genomics applications. Both the input data[56,104,129,168,170] and biological knowledge[113,114,119,126] can be represented as graphs. Edges can encode relationship or similarity between biological entities, such as pathways, functional categorizations, motif gene sets, chromosomal position, and edit distance, while nodes can accommodate observed variables, such as gene expressions, as well as unobserved higher-level entities, such as proteins[168,170]. Graph-based models provide a unique opportunity to concurrently explore multiple disease-disease, drug-drug, and drug-disease relationships. This capability enables a shift away from the traditional "one-disease, one drug" paradigm, potentially expediting drug discovery processes. A representative work by Cheng et al.[131] built a network of nearly two thousand clinically approved drugs. Drug-target binding profiles were pooled from multiple data sources and the proximity between two drug's targets were measured based on the interactome between the targets of each drug. Once the network was built, the topological relationship between two drug-target modules reflect whether drugs are pharmacologically distinct, complementary, indirectly similar, similar, or independent[131].

Biological knowledge can be infused into graph-based models in several ways. Graphical models such as Graphical Convolutional Networks (GCN) can seamlessly take graphs formed from data and from knowledge as input and combine them for prediction[56,128–130]. Edges in a graph can be pre-set based on domain knowledge or softly regularize the estimation process. In cases where a subset of edges is known to be important but the value of their importance is unknown, Graph Attention Networks (GAN) can be used to encourage the model to attend to the important edges. A representative example of using GCN[130] is shown in Figure 3.

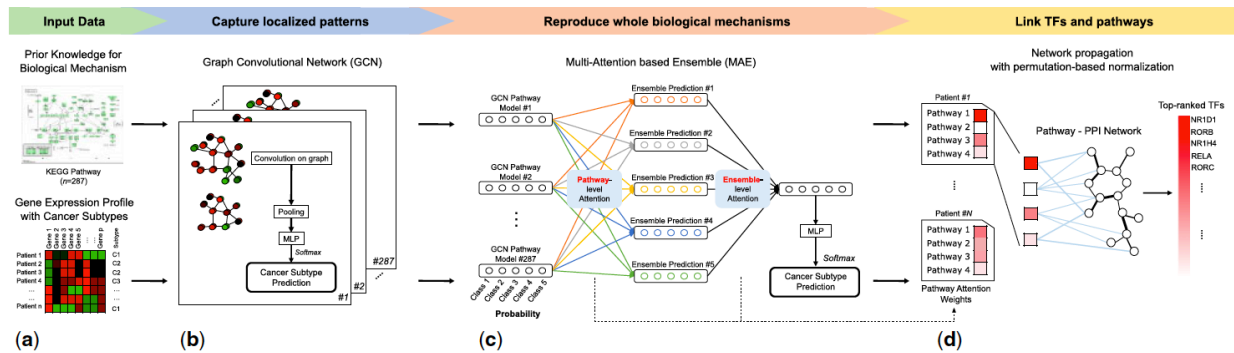

Figure 3. **Workflow of an exemplar GCN model**[130]. (a) Pathways obtained from the KEGG database are used to construct pathway graphs. Gene expression profiles are also encoded in the pathway graphs. (b) One Graph Convolution Network (GCN) is trained for each pathway model to capture localized gene expression patterns. A Multilayer Perceptron (MLP) layer followed by GCN encodes the extracted gene-level information into pathway-level representation. (c) Multi-Attention Ensemble combines the pathway-level representations into sample-level representations. Pathway-level attention highlight pathways most predictive of cancer subtypes. (d) Network propagation on a pathway to protein-protein-interaction (PPI) network is performed to identify transcription factors (TFs) that relate to the pathways highlighted in (c).

Directed graphical models offer a distinctive advantage in their capacity for causal inference, which proves invaluable for comprehending causal relationships between biological entities. One example of this approach is a model designed to propose gene/pathway candidates that potentially enable tumor cell invasion and migration[127]. This model is based on a network with nodes representing biochemical species or phenotypes and edges denoting activations or inhibitory influences. A predefined set of logic rules determines whether a state is reachable from a node. The network was simulated as a Markov decision process to derive probabilities of each node reaching a phenotype.

### Biologically informed neural network architectures

Typically, deep neural networks are known as "black-box" machines with an extensively large number of layers and "neurons". To improve the interpretability of these large neural networks, researchers have proposed to design customized architectures based on biological knowledge, where each layer and node have a specific biological interpretation, and only connections that follow known biological relationships are activated[98,101,102,104,106–111]. A representative example is P-NET[101]: the input layer assimilates the molecular profile of patients as features; the second layer encapsulates a set of genes of interest chosen on the basis of domain knowledge; the subsequent five layers correspond to biological pathways and processes curated from the Reactome dataset (Figure 4). Each node in the second layer is linked to precisely three nodes in the first layer, symbolizing mutations, copy number amplifications, and copy number deletions of each gene. Connections between layers are constructed according to known parent-child relationships based on biomedical knowledge. The customized connections are implemented by multiplying a masking matrix to the weight matrix to nullify undesired connections.

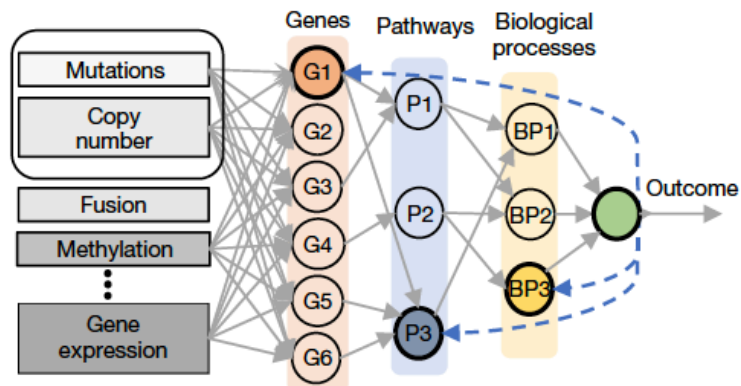

Figure 4. **Schematic overview of the biologically-informed P-NET model**. The input layer takes in molecular data of patients as features. The second layer are defined as genes of interest chosen based on domain knowledge. The next five layers represent pathways and biological processes curated from the Reactome dataset. Each node in the second layer is linked to precisely three nodes in the first layer, symbolizing mutations, copy number amplifications, and copy number deletions of each gene. The trained P-NET provides a relative ranking of nodes in each layer that can be used generate biological hypothesis. These hypotheses were then experimentally and clinically validated.

Employing a biologically informed architecture can significantly improve the interpretability of black-box neural networks. Computational efficiency also improves, as the number of parameters is limited to the number of known biological relationships. However, the performance of these architectures is contingent on the quality and comprehensiveness of domain knowledge. The hard sparsity constraint also limits the network's ability to explore unknown biological connections. One strategy involves using a hybrid architecture with both data-driven and knowledge-informed structures. As seen in Seninge et al.[103], a data-driven encoder captures complex patterns from gene expression profiles, while a decoder with a single linear layer nullifies connections between the latent layer and input layer if they do not belong to a known biological abstraction. The decoder is also constrained with positive weights to maintain the explainability of the direction of biological activity.

## Customized model pipelines

Apart from model architectures, the model pipeline design can also follow properties of biological relationships[105,127,134,139,153,165,208] or real clinical decision-making processes[138,151,152,165]. For example, Wang et al.[208] proposed a two-stage model pipeline to accommodate the hierarchical relationship of three gene modules relevant to GBM prognosis. Predictions are first generated for the gene module that has more supervision from biological knowledge. The outputs of this step then provided additional information for the interplay of the other two gene modules. As an example of mimicking the decision-making processes of pathologists in breast cancer diagnosis, Mercan et al.[151] proposed a sequence of binary classification models in which a decision is made for a single diagnosis: invasive vs. non-invasive, then non-invasive samples into preinvasive and benign, and finally preinvasive samples into ductal carcinoma in situ and atypia. The underlying reasoning is that features that describe one type of diagnosis do not apply to other diagnosis.

## Customized model views

The input size and receptive field are important design aspects that directly impact model's capabilities to extract information. Input sizes, which can be modified through cropping, scaling, or patching the original input, should be selected balancing the required context and resolution needed to solve the problem. Similarly, receptive field dictates the area in input space, e.g. number of pixels in an image, that contributes to a single output unit. Domain experts can visually assess the selection of input size and receptive field as a sanity check for whether the model design suits their problem[213].

Further, it is not expected that every part of an image contributes equally to the diagnosis, and the behavior of radiologists or pathologists when examining an image can offer valuable insights to guide the model where to "look". A simple example is that the high-level organization of tissue at a whole-slide view is crucial for diagnosing invasive cancer, while the cellular features distinguishing preinvasive lesions are often observed at higher resolutions in a local region[151]. Various works attempted to mimic this behavior using multi-view or multi-scale deep learning architectures. Sha et al.[154] developed a multi-field-of-view (FOV) deep learning model to simulate the behavior where pathologists rely on various zoom levels when diagnosing a WSI. The model includes one main network with multiple layers for processing the large image (large FOV) and two branches with only a few layers to process small cuts of the central region of the image (small FOV). This design ensures that the central region is examined more carefully or contributes more to the classification than the edges. Similarly, in classifying challenging nodules from ultrasound images, Liu et al.[152] observed that radiologists focused on the differences with surrounding tissues because malignant nodules tend to have a blurred and irregular margin due to rapid growth. A triple-branch model was designed to mimic this behavior: the first branch analyses low-level features, the second branch takes in context features, and the third branch extracts margin features.

Attention mechanism is another technique widely used in neural networks to encourage the model to selectively focus on certain parts of the input sequence or image that are most relevant to the prediction and suppress irrelevant regions. Several image-based diagnosis studies have used this technique to emphasize the suspicious or discriminative regions of the image[155,205].

## Model training

Another key component of ML is the training algorithm. The loss function is a critical component because it defines the learning direction of the ML model and quantifies how well the model performs on the given task.

### Knowledge-regularized objective

Regularization can be used to incorporate knowledge into the loss function of ML models. A knowledge-based regularization term can be added to the loss function as:

$$L_{total} = \overbrace{\sum_i L(f(x_i), y_i)}^{data-driven\ loss} + \overbrace{\lambda \sum_i L_k(f(x_i), x_i)}^{knowledge-based\ loss},$$

where $L_k$ measures the inconsistency with prior knowledge, and hyper-parameter $\lambda$ determined the relatively importance of data-driven and knowledge-based loss.

One approach is to drive model predictions to be consistent with knowledge such as simulation results[78,84] and probability atlas[62,92] through a soft regularizer term in the objective[62,78,84,92] or hard constraints[39,61,84]. For example, Wang et al.[78] constrained the difference between the predicted tumor cell density by the proposed model and that simulated from the PI model to be small. Loss function can also penalize the predictive loss of each sample differently. In a classification task using WSI patches[133], the loss function was modified such that a higher penalty is given when the model misclassifies patches that fall into regions that pathologists coarsely annotated as abnormal.

Another approach is to regularize feature coefficients[99], attributions[94], or alignment in the latent space[56,97] based on prior knowledge. A Laplacian regularizer is typically used for knowledge that can be represented in a graphical form, such as protein-protein interaction networks used to encourage similar feature representation for related biological entities[97], patient similarity networks to regularize subgroups of patients to have similar predictions[57], or attribution priors to encourage similar feature contribution from functionally related genes[94]. Another commonly used regularizer is group lasso and its variants, which are used to stimulate biologically related entities to have similar coefficients[99,100].

### Physics-informed neural networks

Physics-informed neural networks (PINNs) are neural networks that encode mathematical equations. PINNs leverage the capability of neural networks as universal approximators and use automatic differentiation to solve a customized physics-based loss. PINNs are increasingly being used to solve PDEs that are notoriously difficult to solve using standard numerical approaches.

Consider a PDE parameterized by $\lambda$ for the solution $u(\boldsymbol{x})$:

$$F(x_1, \dots x_n; u_{x_1,\dots,x_n}; \lambda) = 0 \text{ with } \mathcal{B}(\boldsymbol{u}, \boldsymbol{x}) = 0,$$

where $F$ is the governing equation, $\mathcal{B}$ are the boundary conditions, and $u = u(x_1, \ldots, x_n)$ is the unknown. Solving PDE is to find the $u$ function satisfying the equations. PINNs approximate the solution by training a neural network that minimizes a PDE-based loss, which measures the violation of the governing equation, initial conditions, and boundary conditions:

$$L_{PDE} = \frac{1}{N_{PDE}} \sum_{i=1}^{N_{PDE}} \left\| \hat{u}(x^i, t^i) - F(\hat{u}(x^i, t^i), \lambda) \right\|^2 + \gamma_i \frac{1}{N_{IC}} \sum_{i=1}^{N_{IC}} \left\| \hat{u}(x^i, t^i) - u_{IC,i} \right\|^2 + \gamma_b \frac{1}{N_{BC}} \sum_{i=1}^{N_{BC}} \left\| \hat{u}(x^i, t^i) - u_{BC,i} \right\|^2,$$

where $N_{PDE}, N_{IC}, N_{BC}$ are the number samples generated for the PDE residual, initial conditions, and boundary conditions, $u_{IC,i}$ and $u_{BC,i}$ are the initial condition and boundary condition for the system evaluated at $(x^i, t^i)$, and $\gamma_i, \gamma_b$ are weighting hyperparameters. The input to the PINNs includes pairs of system state and solution values sampled at a set of locations called colocation points. Thus, PINNs are capable to train with no ground truth. If experimental data is available, an additional supervised loss term can be added as the data loss.

While the intravoxel incoherent motion (IVIM) model for diffusion-weighted imaging (DWI) have great potential for estimating prognostic cancer imaging biomarkers, it is rarely used clinical because of long fitting time[83]. A fully connected neural network of only three layers and a physics-based loss was able to solve IVIM equally or sometimes more accurately than traditional Bayesian solvers[83]. Likewise, Perez Raya et al.[89] used PINNs to learn the parameters of the Penne's bioheat equation, which was used to detect tumor location based on the breast surface temperature and thermal properties of the tumor. Another innovative use case of PINNs is estimating patient-specific parameters for the PI model, which additionally provided capabilities to visualize forecasts of the intermediate GBM progression curve[90].

## Multi-task and meta-learning

Multi-task learning is trained to predict multiple outcomes at the same time, which is commonly used in drug response prediction. The multi-task setup allows gathering evidence from multiple drugs to find predictive genes, i.e. supervision can be shared across different tasks. If prior knowledge about task dependencies exists, such knowledge can be ejected into the model through a task relation graph[214]. For example, Ammad-ud-din et al.[122] designed a multi-view and multi-task Bayesian Multiple Kernel Learning that can learn shared evidence across drugs without assuming that the same views need to be relevant to all drugs.

Meta-learning is a generalization of multi-task learning where the model is first trained on a distribution of related tasks before being fine-tuned and evaluated on target tasks. This framework allows incorporating both auxiliary datasets and related tasks at the same time. Cho et al.[167] proposed a meta-learning approach that uses various combinations of integrated omics datasets to train a prediction model of cancer survival. Results showed that this meta-learning design allowed finding robust functional and semantic relationships among genes.

## Knowledge-in-the-loop Learning

Experts can actively participate in the model training process by offering human-in-the-loop assessments. For instance, two experienced radiation oncologists blind-reviewed and selected preferred plans among predicted treatment plans for patients with head and neck cancer in[68]. In another study[142], a one-shot active learning approach was employed, wherein pathologists identified the most representative samples from high-fidelity samples screened by a self-distillation model. These selected samples were then labeled as clean data for noise-robust training. One-shot active learning was used where pathologists picked up the most representative samples from high-fidelity samples screened from a self-distillation model and tagged them as clean data for noise-robust training.

Domain knowledge can also be used to dynamically adjust or update model outputs during the training process. For example, Huang et al.[92] first applied a deep learning segmentation model to unlabeled CT scans. The segmentation outputs were compared with a probability atlas to compute a confidence map. The high confidence pixels were then selected and used as pseudo-labels for unlabeled samples. Similarly, Li et al.[93] refined their segmentation maps by warping with the probabilistic atlas map and computing the nearest interpolation.

Reinforcement Learning (RL) builds an environment where biomedical knowledge and data-driven algorithms can seamlessly interact. RL is a ML paradigm where an agent learns optimal decision-making strategies by interacting with an environment and receiving feedback signals in the form of rewards. RL has been used in various clinical applications to control treatment doses, including cancer chemotherapy[58,77]. In a RL framework designed to mimic cancer antiangiogenic therapy, a controller learns optimal drug administration schedules by interacting with a patient whose tumor volume is simulated via a mathematical model of tumor growth[58]. Another representative work leveraged deep reinforcement learning and multiscale PDE to simulate tumor growth, where biological knowledge played an essential role in defining the regulatory substances, diffusible factors, cancer cells, and signaling pathways as RL entities in this virtual tumor microenvironment[75].

### Hyperparameter Setting

Domain knowledge can also be used to set model hyperparameters. In tumor detection and segmentation models, prior knowledge about tumor size and shapes have been used to define realistic anchor boxes[138,152]. Radiologists' sparse annotations of prostate contour have been used as prior seed points for a ML model to produce semi-automatic segmentations[140,141]. Bayesian models such as Bayesian Networks can utilize domain knowledge to define appropriate priors[96]. Expectations about smoothness of predicted tumor cell density maps or sparsity of predictive gene sets can also be used to determine appropriate ranges for the corresponding hyperparameters.

## Challenges and Future Perspectives

A primary limitation in developing accurate and generalizable ML models for cancer diagnosis and prognosis is the scarcity of labeled data. Despite ongoing efforts in data collection, obtaining large-scale clinical samples remains challenging and costly due to factors like invasive biopsy procedures and the substantial time needed for expert annotation. Integrating biomedical knowledge with ML models serves to partially mitigate the shortage of labeled data, enhancing model effectiveness while grounding predictions in established biomedical understanding. As ML becomes increasingly popular in cancer research, there is a growing need for future research to explore diverse and innovative KIML techniques, taking into account the specific nature and characteristics of knowledge relevant to each task.

Another key benefit of KIML lies in improving the explainability of 'black box machines', particularly deep learning models. For real clinical usage, the ability for clinicians to understand and justify the predictions made by AI models is paramount. Explainable models provide insights into the rationale behind predictions, enabling healthcare professionals to assess the reliability and validity of these predictions in the context of patient care. More importantly, explainable models can contribute to a deeper understanding of cancer's complex mechanisms. By comprehensively exploiting patterns from large, multi-modal datasets using AI, Researchers can uncover new hypotheses or validate existing ones about cancer biology. Future research should prioritize and validate the explainability of their models, considering approaches such as coupling KIML with XAI tools like SHapley Additive exPlanations (SHAP)[215].

Moreover, the incorporation of active learning strategies, which iteratively updates the model based on successive treatment outcomes, can significantly tailor and refine the model's performance over time. This adaptive approach is not only able to optimize sample efficiency but also ensure its improvement, aligning closely with the evolving nature of clinical treatments and patient responses. Traditional active learning involves human experts to provide feedback during the process based on their clinical assessment or experimental evaluation. This concept can be generalized as knowledge-in-the-loop learning, where the feedback system may consist of domain experts or an automated oracle constructed based on existing scientific knowledge.

In addition, collecting large-scale public datasets for different types of tumors and drug has greatly fostered collaboration, transparency, benchmarking, and knowledge discovery in cancer diagnosis and prognosis. Public datasets facilitate the pre-training of robust models, which can be fine-tuned with smaller, specific datasets, through techniques such as transfer learning, partially mitigating the shortage of limited samples. Moreover, future research that concentrate on compiling a collective band of quantitative biomedical knowledge can greatly facilitate knowledge integration into ML models.

Several design considerations arise when designing KIML models. First, there is an inherent trade-off between knowledge exploration and exploitation based on whether the modeling goal is knowledge-discovery or consistency. Different types of knowledge integration techniques are suitable for each case. For instance, biologically informed architectures constrain the model to explore relationships among known biological entities, while other techniques, such as knowledge-based regularizers, dynamically allow the model to disobey knowledge and explore new possibilities. Second, knowledge selection is not trivial, especially in unexplored problems like discovering the effect of a new drug or exploring mechanisms of unknown biological phenomenon. A systematic way to probing and selecting relevant knowledge from existing knowledge banks is needed. Third, similar to experimental data, knowledge can contain subjectivity and selection bias. Therefore, the ability to quantify uncertainty in knowledge and dynamically adjust the model's response is desired. There is ample room for future improvements in KIML.

Ultimately, the central aim is to develop user-centric AI systems to advance cancer diagnosis, prognosis, and treatment. KIML is a nonnegligible component to enhance model interpretability and reliability in building AI systems for medical decision-making[197]. In the development of such systems, it is crucial to involve clinical users who provide feedback about what domain knowledge is accessible and what model explanations are relevant to their routine clinical workflow[197].

# Conclusions

With the advancements of AI, ML has gained increasing attention in healthcare research, emerging as potent tool for the analysis of diverse data modalities such as radiologic images, electronic health records, histopathologic slides, and molecular profiles. In cancer applications, unique modeling challenges arise, such as inter-patient and intratumoral heterogeneity, rendering one-fits-all, off-the-shelf ML models less effective. Integrating existing biomedical knowledge is a promising strategy to provide guidance to data-driven models with the potential to enhance performance, generalizability, and interpretability. The key design questions of which knowledge to utilize, how to represent it, and how to integrate it into models necessitate customized solutions on a case-by-case basis. Despite the wealth of available biomedical knowledge and vast datasets harnessed by the biomedical community, much remains unexplored. Collaboration between ML scientists and medical experts will be critical to develop better data-driven and knowledge-informed analytical tools, poised to advance the realms of diagnostics, prognostics, and knowledge discovery of cancer.

# Acknowledgements

This work was supported by NIH U01CA250481-01A1, and NSF DMS-2053170.

# Author contributions

L.M., H.W., and J.L. designed the scope and structure of the Review. L.M. and H.W. collected data, categorized articles, and wrote the manuscript. K.R.S. wrote the sections related to mathematical models. All authors contributed substantially to discussion of the content and reviewed the manuscript before submission.

# Competing interests

The authors declare no competing interests.

# Data availability

We host a live table summarizing the KIML methods in https://lingchm.github.io/kinformed-machine-learning-cancer/. This table contains all papers included in this review and will be updated with new works in the literature, providing an evolving resource for the community.